# Dimension Correction for Hierarchical Latent Class Models

Tomáš Kočka
Department of Computer Science
Aalborg University, Denmark
kocka@cs.auc.dk

Nevin L. Zhang *
Department of Computer Science
Aalborg University, Denmark
nevin@cs.auc.dk

## Abstract

Model complexity is an important factor to consider when selecting among graphical models. When all variables are observed, the complexity of a model can be measured by its standard dimension, i.e. the number of independent parameters. When hidden variables are present, however, standard dimension might no longer be appropriate. One should instead use effective dimension (Geiger *et al.* 1996). This paper is concerned with the computation of effective dimension. First we present an upper bound on the effective dimension of a latent class (LC) model. This bound is tight and its computation is easy. We then consider a generalization of LC models called hierarchical latent class (HLC) models (Zhang 2002). We show that the effective dimension of an HLC model can be obtained from the effective dimensions of some related LC models. We also demonstrate empirically that using effective dimension in place of standard dimension improves the quality of models learned from data.

## 1 INTRODUCTION

Learning graphical models from data has been widely studied in recent years. Two aproaches to learning have been developed: one uses independence tests to search among models and the other uses a score to search for the best model - a procedure known as *model selection*.

Cooper & Herskovits (1992) derived a formula for computing the exact Bayesian score ($p(D|G)$, *marginal likelihood* of a model structure $G$ given data $D$) in the case of complete data and showed that exact computation of the score is intractable when hidden variables are present. In such a case asymptotic approximations of the marginal likelihood such as *Bayesian Information Criterion* (BIC) (Schwarz 1978) and *Cheeseman-Stutz Criterion* (CS) (Cheeseman & Stutz 1995) are usually employed.

The BIC score has two parts: one evaluates the fit of the model to the data and the other penalizes the model according to its dimension. The standard dimension might not be correct when hidden variables are present. Consider the model $O \to X$ with two variables - observed $O$ and hidden $X$. All the parameters in $P(X|O)$ are irrelevant as they do not influence the fit of the model to the (observed) data. Thus there is no reason to penalize the model for such parameters.

Geiger *et al.* (1996) introduced the *effective dimension* for models with hidden variables and related it to the rank of the Jacobian matrix of the transformation between the parameters of the model and the parameters of the distribution over the observed variables. They modified the BIC and CS scores by accounting for the dimension correction. They computed the rank numerically for some models and conjectured that the differences between the standard and effective dimension are rare for LC models (they found just one such model).

Settimi & Smith (1998, 1999) studied effective dimension for the special case of trees with binary variables and for the special case of two observed and one hidden variable. They fully described these two special cases.

In this paper we first study the effective dimension of LC models. We present many LC models in which the standard and effective dimensions differ. We introduce three natural upper bounds and show that the minimum of these is a tight upper bound approximation. We discuss in which situations which upper bound ap-plies. We have found only two LC models for which the effective dimension is not equal to the upper bound derived - in both cases the bound overestimates the number of effective parameters by one.

We then study the effective dimension of HLC models which generalize the LC models by enabling local dependencies among the observed variables. We show that the true number of effective parameters of an HLC model can be computed, by use of a simple rule, from the number of effective parameters of some LC models which are local parts of the HLC model.

Most researchers (e.g. Chickering & Heckerman 1997, Zhang 2002) leave the dimension correction out of the learning. We empirically demonstrate that accounting for the dimension correction leads to better approximation of the probability distribution over the observed variables for LC models. However, dimension correction applies to only few LC models of practical interest. We show that the better approximation is observed for HLC models as well and it concerns many HLC models of practical interest.

## 2 BASIC CONCEPTS

In this section we review basic concepts of graphs, graphical models, latent class models, scores used for model selection and known results concerning effective dimension of models with hidden variables.

### 2.1 GRAPHS AND GRAPHICAL MODELS

A *graph* $G$ is a pair $(N, E)$, where $N$ is a set of nodes and $E$ is a set of edges, i.e. subset of $N \times N$ of ordered pairs of distinct nodes. Each node $X \in N$, denoted by an upper-case letter, represents a discrete variable. We denote the number of states of a variable $X$ by $|X|$ and a particular state of a variable $X$ by a lower-case letter $x$. We often use a set of variables $R \subseteq N$ to represent a joint variable over its elements which has number of states $|R| = \prod_{X \in R} |X|$.

An *Acyclic Directed Graph* (DAG) is a graph where all edges are directed and there are no cycles. If a graph has directed edge $A \to B$, then the node $A$ is *parent* of the node $B$, i.e. $A \in Pa(B)$, and $B$ is *child* of $A$, i.e. $B \in Ch(A)$. The union of children and parents of a node is called *neighbours*, i.e. $Ne(A) = Pa(A) \cup Ch(A)$.

A *tree* is a connected undirected graph without cycles. A *directed tree* is a DAG obtained from a tree by choosing a root node and directing all edges away from this node. A tree has one edge less than the number of nodes. It has a unique path between any two vertices. We say that two sets of nodes $R, T \in N$ are *separated* by $S \in N$ in a graph $G$ if every path from $R$ to $T$ in $G$ contains a node from $S$.

A *Bayesian network* is a pair $(G, \theta_G)$ where $G$ is a DAG and $\theta_G$ are parameters. The parameters are conditional probabilities for each node $X \in N$ given its parents $Pa(X)$, i.e. $P(X|Pa(X))$. The standard number of (free, independent) parameters $ds$ in a Bayesian network is

$$ds(G) = \sum_{X \in N} (|X| - 1) * \prod_{Y \in Pa(X)} |Y|.$$

A Bayesian network represents a joint probability distribution $P(N|G, \theta_G)$ over all variables $N$ using the *factorization* formula

$$p(N|G, \theta_G) = \prod_{X \in N} p(X|pa(X)).$$

A node $A$ in a tree is separated by its neighbours $Ne(A)$ from all other nodes. Thus each distribution, which factorizes according to a tree, satisfies the conditional independence $N \setminus A \setminus Ne(A) \perp\!\!\!\perp A | Ne(A)$, i.e. $P(N) * P(Ne(A)) = P(N \setminus A) * P(A \cup Ne(A))$.

We say that two graphical models are *equivalent* if they represent the same class of probability distributions over all observed variables.

### 2.2 LC AND HLC MODELS

A *Latent Class* (LC) model (see Figure 1) is a graphical model with one hidden variable $X$ and observed variables $O$. It can be represented as a directed or undirected tree where the observed nodes coincide with nodes having exactly one neighbour. We refer to instances of the LC model by $|X| : |O_1|, ..., |O_i|, ...|O_n|$.

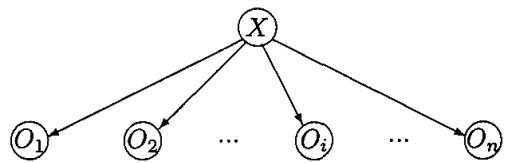

Figure 1: Latent Class model

A *Hierarchical Latent Class* (HLC) model (see Figure 2) is a graphical model with observed nodes $O$ and hidden nodes $H$. HLC models are an extension of LC models introduced in (Zhang 2002) which enable modelling of local dependencies (dependencies among subsets of observed variables). An HLC model can be represented as a directed or undirected tree where the observed nodes coincide with nodes having exactly one neighbour. We refer to instances of the particular HLC model in Figure 2 by $|H_1|, |H_2|, |H_3| : |O_1|, ..., |O_5|$.

Each hidden node $H_i$ in an HLC model induces, together with its neighbours $Ne(H_i)$, a *local LC model*. It has hidden variable $H_i$ and observed variables $Ne(H_i)$.

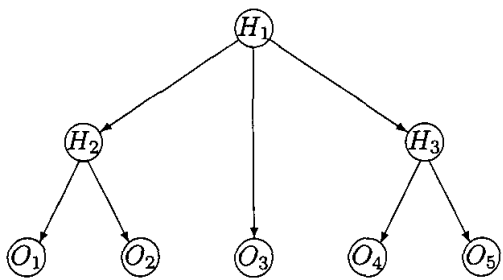

Figure 2: Hierarchical Latent Class model

## 2.3　SCORES FOR MODEL SELECTION

All the scores are approximations of the marginal log likelihood $\log p(D|G)$ of data $D$ given the graphical model structure $G$. They use the *maximum likelihood* (ML) or *maximum aposterior probability* (MAP) estimate $\hat{\theta}_G$ of the model parameters $\theta_G$. Maximum likelihood estimates in the case of missing data or hidden variables can be obtained by use of EM algorithm or gradient descent methods. Some scores use the completed data $\hat{D}$ obtained from $D$ using $\hat{\theta}_G$. Note that the $p(\hat{D}|G)$ can be evaluated by use of the formula for exact marginal likelihood. We denote by $de(G)$ the *effective dimension* of the model with hidden variables. Note that the standard dimension of the model and the effective dimension of the model with all variables observed are both $ds(G)$.

The BIC score in (Schwartz 1978), the CS score in (Cheeseman & Stutz 1995), the CS and BIC scores with dimension correction in (Geiger et al. 1996) are

$$BIC(D|G) \approx \log p(D|G, \hat{\theta}_G) - ds(G)/2 * \log |D|$$

$$CS(D|G) \approx \log p(\hat{D}|G) - \log p(\hat{D}|G, \hat{\theta}_G) + \log p(D|G, \hat{\theta}_G)$$

$$CS_+(D|G) \approx CS(D|G) + (ds(G) - de(G))/2 * \log |D|$$

$$BIC_+(D|G) \approx \log p(D|G, \hat{\theta}_G) - de(G)/2 * \log |D|.$$

The scores with dimension correction were never used in practice because there are no methods for computing the effective dimension $de$.

## 2.4　EFFECTIVE DIMENSION

A graphical model $G$ transforms its parameters $\theta_G$ into a probability distribution $P(O)$ over all observed variables $O$ (marginal of $P(N)$). We will denote by $J_O(\theta_G) = [J_{jk}] = [\frac{\partial p(o_j)}{\partial \theta_k}]$ the Jacobian matrix of this transformation. Rows of $J_O(\theta_G)$ correspond to states in the observed space $O$ of the model $G$ and columns to the parameters $\theta_G$. Geiger et al. (1996) showed that the *effective dimension $de(G)$* of a model $G$ is the rank of $J_O(\theta_G)$. The rank in general is a function of $\theta_G$ but was shown to be constant almost everywhere. We use $J_O(\theta); \theta \subseteq \theta_G$ to denote a matrix which has only a subset of columns in $J_O(\theta_G)$.

This suggests the following numerical approach to computing $de$: generate random $\theta$, compute the Jacobian and its rank with sufficient numerical precision. We performed this computation for many LC and HLC models in Maple. We repeated each computation ten times in our experiments and we corroborate the observation in (Geiger et al. 1996) that none of the randomly chosen parameters $\theta$ accidentally reduced the rank.

The *rank* of a matrix is a number of (row or column) vectors in a basis of the matrix. A *basis* is a set of linearly independent vectors such that all other vectors can be expressed as a linear combination of the vectors in the basis. Thus

$$de \leq ds \text{ and } de \leq dc$$

where $dc$ is the number of parameters in the complete model over all observed variables $O$

$$dc = \prod_{X \in O} |X| - 1.$$

Moreover there are two special cases for which theorethical solution for $de$ is known.

THEOREM 1 *(Settimi & Smith 1998)*
*The LC model $O_1 \leftarrow X \rightarrow O_2$ where $|X| \leq \min(|O_1|, |O_2|)$ has $|X| * (|X| - 1)$ unidentifiable parameters, i.e. $de = ds - |X| * (|X| - 1)$. If $|X| \geq \min(|O_1|, |O_2|)$, then the hidden variable does not impose any restriction on the observed marginal $P(O_1, O_2)$ and thus $de = dc$.*

THEOREM 2 *(Settimi & Smith 1999)*
*An HLC model with all variables binary and $k$ hidden nodes with less than three neighbours has $2 * k$ unidentifiable parameters, i.e. $de = ds - 2 * k$.*

## 3　DIMENSION OF LC MODELS

We have already seen two general upper bounds on $de$, namely $ds$ and $dc$. In this section we introduce another upper bound for LC models. We combine all these into one upper bound and show that it is a very tight upper bound for LC models. Then we show how to reduce the space of all LC models without changing its modelling power.

## 3.1 UPPER BOUNDS

THEOREM 3
*Let $M$ be an LC model with observed variables $O$ and hidden variable $X$. Let $M^*$ be another LC model with two observed variables $U_1, U_2$ and hidden variable $X$. Let $O' \subset O$, $U_1 = O'$ and $U_2 = O \setminus O'$. If $|X| < min(|U_1|, |U_2|)$, then $de(M) \leq de(M^*) = ds(M^*) - |X| * (|X| - 1)$, else $de(M) \leq de(M^*) = dc(M^*) = dc(M)$.*

**Proof:** The two LC models $M$ and $M^*$ have the same joint observed space $O$ and the same hidden variable. Any probability distribution over $O$ represented by the model $M$ can be represented by the model $M^*$ as well. $de(M) \leq de(M^*)$ because the model $M$ applies some additional constraints in comparison with the model $M^*$. The rest follows directly from Theorem 1. □

This result introduces a whole class of upper bound limits. We denote the lowest one by

$$dp(M) = min_{M^*} de(M^*) \quad ;|X| < min(|U_1|, |U_2|)$$
$$= dc(M) \text{ otehrwise.}$$

By combining the upper bounds we get the following theorem.

THEOREM 4
*For any LC model M,*

$$de(M) \leq db(M) = min(ds(M), dc(M), dp(M)).$$

**Proof:** It follows from the definition of $de$ and Theorem 3. □

The next lemma states when $db(M) = ds(M)$ and thus simplifies the computation of $db(M)$.

LEMMA 3.1
*For an LC model with observed variables $O$ and hidden variable $X$, if $|X| < 2 * \sqrt{|O|} - \sum_{i=1}^{n} |O_i| + (n-1)$ and $|X| < \frac{|O|}{\sum_{i=1}^{n} |O_i| - (n-1)}$, then $ds < dp$ and $ds < dc$.*

**Proof:** Follows directly from the definition of $ds$, $dc$ and $dp$ togehter with the fact that $dp \geq |X| * (2 * \sqrt{|O|} - |X|) - 1$ as $|U_1| + |U_2| \geq 2 * \sqrt{|O|}$. □

We can see from Lemma 3.1 that for many observed variables and a reasonably small number of states of all variables, the standard dimension $ds$ applies. However, for models with few observed variables (see Table 1) this is not the case. There are many models where $de \neq ds$. Table 1 suggests that the upper bound $db$ from Theorem 4 is tight. We have found only two LC models (3:2,2,2,2 and 4:3,3,3) for which the upper bound $db$ overestimates the true $de$ by one. Note that all three bounds $ds, dc, dp$ apply when evaluating $db$.

Table 1: Comparison of the effective dimension $de$ computed numerically in Maple, bound $db$ from Theorem 4 (bold if $db \neq de$), standard dimension $ds$, complete dimension $dc$ and pairwise dimension $dp$ from Theorem 3 (bold if $ds, dc, dp = db$) for selected LC models (see Figure 1).

| LC model | de | db | ds | dc | dp |
|---|---|---|---|---|---|
| 2:2,2 | 3 | 3 | 5 | **3** | dc |
| 2:2,2,2 | 7 | 7 | **7** | **7** | dc |
| 3:2,2,2 | 7 | 7 | 11 | **7** | dc |
| 4:2,2,2 | 7 | 7 | 15 | **7** | dc |
| 2:3,3 | 7 | 7 | 9 | 8 | **7** |
| 2:3,3,3 | 13 | 13 | **13** | 26 | 19 |
| 3:3,3,3 | 20 | 20 | **20** | 26 | dc |
| 3:4,5 | 17 | 17 | 23 | 19 | **17** |
| **4:3,3,3** | 25 | **26** | 27 | **26** | dc |
| 5:3,3,3 | 26 | 26 | 34 | **26** | dc |
| 6:3,3,3 | 26 | 26 | 41 | **26** | dc |
| 2:2,2,2,2 | 9 | 9 | **9** | 15 | 11 |
| **3:2,2,2,2** | 13 | **14** | **14** | 15 | **14** |
| 4:2,2,2,2 | 15 | 15 | 19 | **15** | dc |
| 5:2,2,2,2 | 15 | 15 | 24 | **15** | dc |
| 6:2,2,2,2 | 15 | 15 | 29 | **15** | dc |
| 3:5,2,2 | 17 | 17 | 20 | 19 | **17** |
| 3:4,2,2 | 14 | 14 | 17 | 15 | **14** |
| 5:3,3,2 | 17 | 17 | 29 | **17** | dc |
| 5:6,3,2 | 34 | 34 | 44 | 35 | **34** |
| 5:10,3,2 | 54 | 54 | 64 | 59 | **54** |

## 3.2 REGULAR LC MODELS

LEMMA 3.2
*Let $M$ be an LC model with observed variables $O$ and hidden variable $X_1$ where $|X_1| > \frac{|O|}{max_i|O_i|}$. Let $M^*$ be an LC model with observed variables $O$ and hidden variable $X_2$ where $|X_2| = \frac{|O|}{max_i|O_i|}$. Then $M$ and $M^*$ are equivalent models.*

**Proof:** It is obvious that any distribution over $O$ encoded by $M^*$ can be encoded by $M$ as well. We will show that any distribution over $O$ can be encoded by $M^*$. Let $|O_j| = max_i|O_i|$. Assign $P(O \setminus O_j | X_2)$ in such a way that $X_2 = O \setminus O_j$. Then $X_2 \to O_j$ is a complete model which can encode any distribution. □

We say that an LC model is *regular* if $|X| \leq \frac{|O|}{max_i|O_i|}$. Each irregular LC model is equivalent to some regular LC model. Thus, the modelling power of the class of LC models is not reduced if we restrict ourselves to the class of regular LC models.

## 4 DIMENSION OF HLC MODELS

In this section we show how to compute the effective dimension of HLC models. Consider the HLC model 5,3,3:2,2,2,2,2 (see Figure 2). Its standard dimension is 41 while its effective dimension is 23 parameters. The difference between the standard and effective parametrization is 18 parameters. There are three hidden nodes in the HLC model. They induce local LC models 3:5,2,2 ; 5:3,3,2 and 3:5,2,2. The differences between the standard and effective parametrization for these LC models can be read from Table 1. They are 3, 12 and 3. The sum of these differences is 18. This equals the difference for the HLC model. The same rule applies to all HLC models we tested (with different graphical structures). In this section, we prove that this fact is true in general.

LEMMA 4.1
*Let $M$ be an HLC model with observed variables $O$ and hidden variables $H$ where for some $H_i \in H$ holds that $|H_i| > \frac{|Ne(H_i)|}{max_{X \in Ne(H_i)}|X|}$. Let $M^*$ be the same HLC model as $M$ except that $|H_i| = \frac{|Ne(H_i)|}{max_{X \in Ne(H_i)}|X|}$. Then $M$ and $M^*$ are equivalent models.*

**Proof:** The LC models induced by the hidden node $H_i$ (their observed space is $Ne(H_i)$) in $M$ and $M^*$ are equivalent (see Lemma 3.2). In both models $M$ and $M^*$ the independence $H_i \perp\!\!\!\perp N\backslash H_i\backslash Ne(H_i)|Ne(H_i)$ holds. Thus $M$ and $M^*$ are equivalent models as well. □

We say that an HLC model is *regular* if for each hidden node $H_i$ holds that $|H_i| \leq \frac{|Ne(H_i)|}{max_{X \in Ne(H_i)}|X|}$ and the strict inequality holds if $H_i$ has exactly two neighbours $Ne(H_i)$ where $Ne(H_i) \cap H \neq \emptyset$. Each irregular HLC model is equivalent to some regular HLC model.

Our task is to estimate the effective dimension $de$ of a regular HLC model $M$. If $M$ has just one hidden node, then it is an LC model and we can use the results of Section 3. If $M$ has more hidden nodes, then the theorem below enables us to decompose the problem into two smaller HLC models.

THEOREM 5 *Let $M$ be a regular HLC model with observed variables $O$ and hidden variables $H$. Let $X \in H$ be the root node and $Z$ be a hidden child of $X$. Let $N_1$ be the set of nodes separated from $Z$ by $X$ in $M$, and $N_2$ be the set of nodes separated from $X$ by $Z$ in $M$. Let $M_1$ and $M_2$ be the HLC models induced from $M$ by nodes $N_1 \cup \{X,Z\}$ and $N_2 \cup \{X,Z\}$ respectively. Then $M_1$ and $M_2$ are regular HLC models and $ds(M)-de(M) = ds(M_1)-de(M_1)+ds(M_2)-de(M_2)$.*

**Proof:** Figure 3 shows the situation. Note that $N_1 \cup N_2 \cup \{X,Z\} = O \cup H$. Let $O_1 = O \cap N_1$ and $O_2 = O \cap N_2$. Note that $M_1$ has observed variables $O_1 \cup Z$ and hidden variables $H \cap N_1 \cup X$ and $M_2$ has observed variables $O_2 \cup X$ and hidden variables $H \cap N_2 \cup Z$. Let $M_0$ denote the common part of $M_1$ and $M_2$, i.e. the model $X \to Z$. The three Jacobian matrices for models $M$, $M_1$ and $M_2$ are $J_O(\theta_M)$, $J_{O_1 \cup Z}(\theta_{M_1})$ and $J_{O_2 \cup X}(\theta_{M_2})$. Both models $M_1$ and $M_2$ are regular as the regularity conditions are the same in $M$.

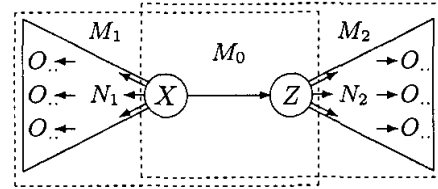

Figure 3: HLC model $M$ and its induced submodels

The proof is organized as follows. First we show that $J_O(\theta_{M_0})$ is basis of its column space. Second we relate $J_O(\theta_{M_1})$ and $J_{O_1 \cup Z}(\theta_{M_1})$, resp. $J_O(\theta_{M_2})$ and $J_{O_2 \cup X}(\theta_{M_2})$. Then we show that $ds(M) - de(M) \geq ds(M_1) - de(M_1) + ds(M_2) - de(M_2)$. Third we show that $ds(M) - de(M) = ds(M_1) - de(M_1) + ds(M_2) - de(M_2)$.

I. The Jacobian $J_{XZ}(\theta_{M_0})$ of the model $M_0$ is obviously basis of itself. We prove that $J_O(\theta_{M_0})$ is basis of its column space by contradiction. Assume that the column vectors in $J_O(\theta_{M_0})$ are linearly dependent for any parameters $\theta_M$. We can choose the parameters $\theta_{M_1} \backslash \theta_{M_0}$ in such a way that for any state $x$ of $X$ there is exactly one unique state $o$ (different $o$ for different $x$) in $O_1$ such that $p(x|o) = p(o|x) = 1$. The same holds for $\theta_{M_2} \backslash \theta_{M_0}$, any state $z$ of $Z$ and $O_2$. This is always possible because for any hidden node $H_i \in H$ in any regular HLC model $|H_i| \leq |Ch(H_i)| \leq \frac{|Ne(H_i)|}{max_{Y \in Ne(H_i)}|Y|}$. Thus, for any state $xz$ of $XZ$ there is exactly one state $o$ of $O$ such that $p(xz|o) = 1$. This means that all rows in $J_{XZ}(\theta_{M_0})$ are in $J_O(\theta_{M_0})$ which contradicts the fact that $J_{XZ}(\theta_{M_0})$ is basis of itself.

II. For any $\theta_i \in \theta_{M_1}$ holds that

$$J_O(\theta_i) = \sum_Z P(O_2|Z) * J_{O_1 \cup Z}(\theta_i)$$

because $O_2 \perp\!\!\!\perp O_1|Z$, $O_1 \cup O_2 = O$ and $P(O_2|Z)$ does not depend on $\theta_{M_1}$. The same holds for $J_O(\theta_{M_2})$ and $J_{O_2 \cup X}(\theta_{M_2})$. Thus, any set of linearly dependent column vectors in $M_1$ ($M_2$) is linearly dependent in $M$ as well. Consequently, columns of $J_{O_1 \cup Z}(\theta_{M_0})$ are linearly independent because columns of $J_O(\theta_{M_0})$ are linearly independent. The same holds for $J_{O_2 \cup X}(\theta_{M_0})$. Thus, there exists a basis $J_{O_1 \cup Z}(\theta_{B_1})$ of the column space of $J_{O_1 \cup Z}(\theta_{M_1})$ where $\theta_{M_0} \subseteq \theta_{B_1} \subseteq \theta_{M_1}$. Similarly exists $J_{O_2 \cup X}(\theta_{B_2})$ where $\theta_{M_0} \subseteq \theta_{B_2} \subseteq \theta_{M_2}$.

Then, there exists a basis $J_O(\theta_B)$ of the column space of $J_O(\theta_M)$ where $\theta_B \subseteq \theta_{B_1} \cup \theta_{B_2}$. Thus, $|\theta_B| \leq |\theta_{B_1}| + |\theta_{B_2}| - |\theta_{M_0}|$. This is equivalent to $ds(M) - de(M) \geq ds(M_1) - de(M_1) + ds(M_2) - de(M_2)$ because $de(M) = |\theta_B|$, $de(M_1) = |\theta_{B_1}|$, $de(M_2) = |\theta_{B_2}|$ and $|\theta_{M_0}| = ds(M_1) + ds(M_2) - ds(M)$.

**III.** We prove that $J_O(\theta_{B_1} \cup \theta_{B_2})$ is basis of its column space, i.e. $\theta_B = \theta_{B_1} \cup \theta_{B_2}$, by contradiction. Assume that the columns in $J_O(\theta_{B_1} \cup \theta_{B_2})$ are linearly dependent for any $\theta_M$ and denote by $k_i$ the weight of each column corresponding to the parameter $\theta_i \in \theta_{B_1} \cup \theta_{B_2}$ in the linear combination which yields a zero vector. We can choose the parameters $\theta_{M_1} \setminus \theta_{M_0}$ in such a way that for any state $x$ of $X$ there is exactly one unique state $o_x$ (different $o_x$ for different $x$) in $O_1$ such that $p(x|o_x) = p(o_x|x) = 1$. Denote by $J_{o_x,O_2}(\theta_{B_1} \cup \theta_{B_2})$ the matrix having all rows from $J_O(\theta_{B_1} \cup \theta_{B_2})$ corresponding to the state $o_x$. The linear combination of columns in $J_{o_x,O_2}(\theta_{B_1})$ is $\sum_{\theta_i \in \theta_{B_1}} k_i * J_{o_x,O_2}(\theta_i) = \sum_{\theta_i \in \theta_{B_1}} k_i * \sum_Z P(O_2|Z) * J_{o_x,Z}(\theta_i) = \sum_Z P(O_2|Z) \sum_{\theta_i \in \theta_{B_1}} k_i * J_{o_x,Z}(\theta_i)$. There are always weights $k_j^*$ for $\theta_j \in (\theta_{Z|x} \cup \theta_x) \subset \theta_{M_0} \subset \theta_{B_2}$ which give the same linear combination as $\sum_{\theta_i \in \theta_{B_1}} k_i * J_{o_x,Z}(\theta_i)$.

We started with linear dependence of columns in $J_{o_x,O_2}(\theta_{B_1} \cup \theta_{B_2})$ and we showed new linear dependence of columns in $J_{o_x,O_2}(\theta_{Z|x} \cup \theta_x \cup (\theta_{B_2} \setminus \theta_{M_0}))$ for each state $x$ in $X$. Because $\theta_{Z|x} \cup \theta_x \cup (\theta_{B_2} \setminus \theta_{M_0})$ do not influence the relation between $X$ and $O_1$, we can use $J_{o_x,O_2}(\theta_{Z|x} \cup \theta_x \cup (\theta_{B_2} \setminus \theta_{M_0})) = J_{x,O_2}(\theta_{Z|x} \cup \theta_x \cup (\theta_{B_2} \setminus \theta_{M_0}))$. Then we can put together all rows in $J_{x,O_2}(\theta_{Z|x} \cup \theta_x \cup (\theta_{B_2} \setminus \theta_{M_0}))$ for all states $x$ of $X$ and we get linear dependence of columns of $J_{O_2 \cup X}(\theta_{Z|X} \cup \theta_X \cup (\theta_{B_2} \setminus \theta_{M_0}))$. This contradicts the fact that $J_{O_2 \cup X}(\theta_{B_2})$ is basis of its column space. Thus $\theta_B = \theta_{B_1} \cup \theta_{B_2}$ and $ds(M) - de(M) = ds(M_1) - de(M_1) + ds(M_2) - de(M_2)$. □

COROLLARY 4.1
*Let $M$ be a regular HLC model with observed variables $O$ and hidden variables $H$. Let $M_i$ be the local LC model induced by each hidden node $H_i \in H$ in $M$. Then the difference between the number of standard and effective parameters in $M$ is equal to the sum of the differences over all the $M_i$ models, i.e.*

$$ds(M) - de(M) = \sum_{H_i \in H} ds(M_i) - de(M_i).$$

Thus, we can expect differences between $ds$ and $de$ for HLC models even in real domains with many observed variables whenever there is at least one hidden node with few neighbours.

## 5 EXPERIMENTS WITH LC

In this section we experimentally demonstrate that accounting for the effective dimension leads to learning better LC models from data.

We generated ten random parametrizations of each model from the seven regular LC models 2:2,2,2,2 ; 3: ; ... 8:2,2,2,2 with four binary observed variables. We produced five data sets of diferent sizes from each of these parametrizations. We evaluated for each data set all the models using the four scores introduced in Section 2 (we used EM for the ML estimates). We selected for each pair of score and data the best model. In this section we compare the four scores using the average fit of the best model to the true generative distribution (Kullback-Leibler Information divergence) and cardinality of the hidden variable.

Results for data generated from the LC model 8:2,2,2,2 (5: ... 7: are similar) are in Figure 4. The CS and BIC exhibit the same behaviour. They are outperformed by the BIC+ and CS+ in the fit of data for larger samples. The differences in the fit of data between BIC+ and CS+ are not significant. The BIC+ selects higher cardinalities of $X$ than CS+ and CS+ selects higher cardinalities than BIC and CS.

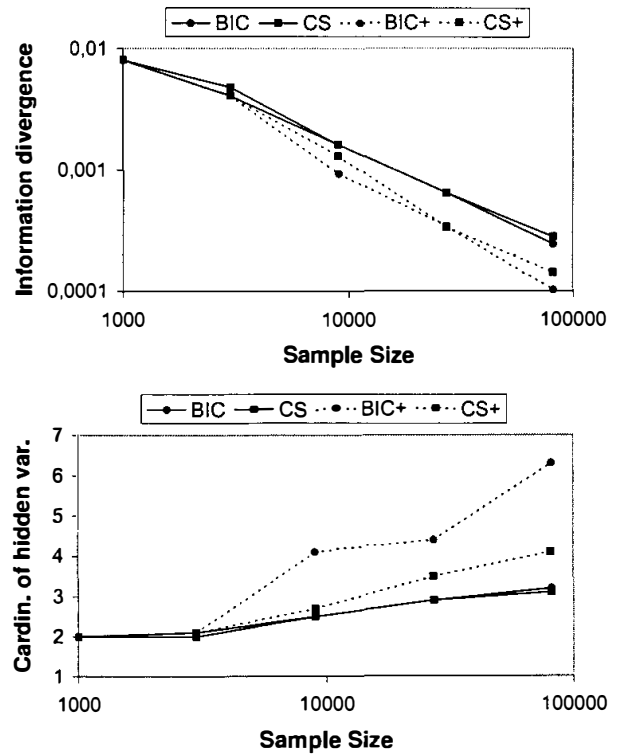

Figure 4: Fit of the generative distribution over $O$ and cardinality of $X$ for LC models learned from data generated from the LC model 8:2,2,2,2 .

For data generated from the LC model 4:2,2,2,2 (3: is similar) all the four scores lead to similar fit. However, the $BIC_+$ gives significantly higher cardinalitites of the hidden variable than the other scores. This behaviour of $BIC_+$ is expected as the likelihood of all the models with cardinality of the hidden variable higher than 4 is the same (in practice, thanks to random fluctuations of the ML estimate, there are some small random differences) and the penalty is the same as well. Thus the $BIC_+$ selects at random among the models 4:2,2,2,2 ... 8:2,2,2,2 .

For data generated from the LC model 2:2,2,2,2 all the four scores behave in a similar way and produce the same results in both the cardinality of the hidden variable and fit of the true distribution.

In general we can say that models with few states of the hidden variable $H$ usually provide pretty good fit of the observed data. Large sample sizes are needed to obtain models with more states of $H$. The $CS_+$ score outperforms the standard CS and BIC scores and leads to higher cardinalities of the hidden variable. $BIC_+$ provides similar fit to the data as $CS_+$ but leads to more states of $H$.

One possible reason why we need large sample sizes to select more complex models is that some of the randomly generated parameters introduce only weak dependencies. Thus, we parametrized the LC model 8:2,2,2,2 by deterministic relation between three observed variables and the hidden variable and by random parametrization of the remaining parameters. Note that such a model can still encode any distribution over the observed variables. With these data we observe the same behaviour as for the 8:2,2,2,2 model, however from much smaller sample sizes (see Figure 5). The plus scores are able to reach the same fit of the true distribution with half the data.

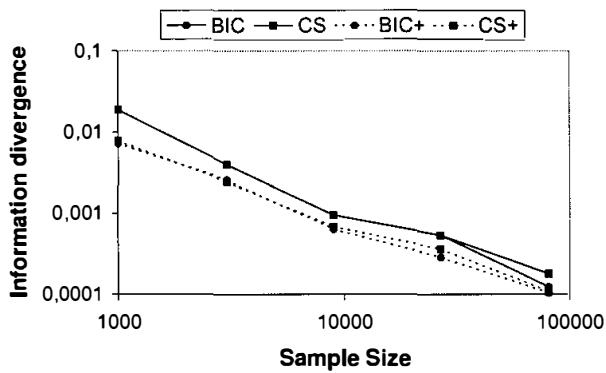

Figure 5: Fit for LC models learned from data generated from a LC model with some deterministic relations.

## 6 EXPERIMENTS WITH HLC

In this section we experimentally demonstrate that accounting for the effective dimension improves the fit of data by HLC models. This fact is of practical importance because for many HLC models $de \neq ds$.

We generated data sets of different sizes from fifty random parametrizations of the HLC model 5,3,3:2,2,2,2,2. This model has 41 standard parameters and 23 effective parameters (see Section 4). It is a regular but not complete model and all its hidden variables have cardinality smaller than the maximal possible according to their neighbours (see Lemma 4.1). Thus, it is a typical example of a model having some hidden node with few neighbours of such cardinalities that it creates a difference between the standard and effective number of parameters.

Zhang (2002) demonstrated that it is usually much easier to recover from the data the true generative structure than the true cardinalities of the hidden variables. Thus we did not score all regular HLC models in our experiments. We considered only the HLC models with the true generative structure and we always started with all hidden variables binary. We applied the hillclimbing aproach to learn the cardinality of the hidden variables, i.e. in each step we increased by one the cardinality for the hidden node where it caused the biggest increase in the score (we again used EM to get the ML estimates). Table 2 summarizes the results of this experiment.

Table 2: Average Information divergence (*E-03 bits) for different scores and sample sizes (SS). The best value for each sample size and all values with no significant difference (95%) from the best one are bold.

| SS | BIC | $BIC_+$ | CS | $CS_+$ |
|---|---|---|---|---|
| 1k | **6.1 ± .6** | 7.5 ± .8 | **6.0 ± .6** | **6.4 ± .6** |
| 3k | **2.2 ± .3** | **2.5 ± .3** | **2.4 ± .3** | **2.4 ± .3** |
| 9k | **.78 ± .1** | **.79 ± .09** | **.76 ± .1** | **.69 ± .07** |
| 27k | .49 ± .08 | **.33 ± .05** | .5 ± .09 | **.38 ± .06** |
| 81k | .22 ± .04 | **.16 ± .03** | .22 ± .03 | **.17 ± .03** |
| 243k | .15 ± .04 | **.1 ± .03** | .15 ± .03 | **.12 ± .03** |

It is clear from Table 2 that using the plus scores which account for the dimension correction leads to better fit of the data. In our experiment this behaviour is observed for larger sample sizes only. This is probably due to the random parametrization of the generative model. We expect that for real data with some deterministic or strong dependencies this behaviour would be observed for smaller sample sizes as well. However, for very large sample sizes all the scores should lead to the same fit of the data, because they have all dimension penalties proportional to the log of the sample size which, compared to the linear proportionality of the likelihood, converge to zero. We never observed this behaviour and we never discovered the generative model, either. The closest model found is 4,3,3:2,2,2,2,2 and it was selected in one run by the $CS_+$ score.

There are some problems with the $BIC_+$ score. The first problem is that for the smallest sample size in Table 2 it resulted in the worst fit. The second problem is that the $BIC_+$ score is not able to discriminate among models with different cardinalities of some hidden variable if they have the same effective number of parameters and if they have the same likelihood (i.e. the simplest one provides the same fit as the others). In fact, the first problem may just be a manifestation of the second problem, as the most frequently learned models for the smallest sample size were 2,2,2: and 2,3,2:, resp. 2,2,3: . The $CS_+$ score deals well with such situations and it is clearly the score of choice according to our experiments.

## 7  CONCLUSIONS

When learning graphical models from data, one typically has to select among multiple models. The BIC score is a popular scoring metric used for this task. The score represents a trade-off between fitness-to-data and model complexity. When all variables are observed, the complexity of a model can be measured by its standard dimension, i.e. the number of independent parameters. Geiger et al. (1996) argue that, when hidden variables are present, the standard dimension might no longer be appropriate. An alternative was proposed. We call it the effective dimension.

A procedure for computing the effective dimension of an LC model is proposed by Geiger et al. (1996). This procedure involves symbolic differentiation and has to be programmed for each model. It is hence difficult to use in practice. Our first contribution in this paper is the alleviation of this difficulty by providing a bound that is tight and easy to compute.

HLC models are a generalization of LC models. They are proposed in (Zhang 2002) to relax the conditional independence assumption of LC models. As the second contribution, we show that the effective dimension of an HLC model can be computed from the effective dimensions of some related LC models. This result applies to any tree with hidden variables.

We have also conducted experiments to gauge the impact of the dimension correction on learning. Our results indicate that dimension correction improves the quality of induced models. In particular, the CS score with dimension correction seems to lead to the best results.

## Acknowledgements

We thank Regitze Larsen, Poul Svante Eriksen, Finn V. Jensen, Jiří Vomlel, Marta Vomlelová, Thomas D. Nielsen, Olav Bangsø, Jose M. Peña, Kristian G. Olesen and the anonymous reviewers for useful remarks. The second author's work on this paper was partially supported by Hong Kong Research Grants Council under grant HKUST6093/99E.